\pdfoutput=1

\documentclass[11pt]{article}

\usepackage[final]{acl}

\usepackage{times}
\usepackage{latexsym}

\usepackage[T1]{fontenc}

\usepackage[utf8]{inputenc}

\usepackage{microtype}

\usepackage{inconsolata}

\usepackage{diagbox}
\usepackage{enumitem}
\usepackage{tikz}
\usetikzlibrary{shapes.geometric, arrows, positioning}
\usepackage{tikzscale}
\usepackage{longtable}
\usepackage{graphicx}
\usepackage{tabularx}
\usepackage{array}
\usepackage{subcaption}
\usepackage{pgfplots}
\usepackage{afterpage}
\usepgfplotslibrary{groupplots}
\usepgfplotslibrary{colorbrewer}

\usepackage{booktabs}
\usepackage{multirow}
\usepackage{xcolor}

\usepackage{ulem} 
\title{``I know myself better, but not really greatly'': How Well Can LLMs Detect and Explain LLM-Generated Texts?}

\author{Jiazhou Ji$^{1\ast}$ \quad Jie Guo$^{1}\thanks{Equal contributions.}$ \quad Weidong Qiu$^{1}$\quad Zheng Huang$^{1}$ \quad Yang Xu$^{1}$ \\\textbf{Xinru Lu$^{1}$ \quad Xiaoyu Jiang$^{1}$ \quad Ruizhe Li$^{2\dag}$\quad Shujun Li$^{3}\thanks{Corresponding co-authors: \texttt{ruizhe.li@abdn.ac.uk, s.j.li@kent.ac.uk}}$} \\
$^1$School of Cyber Science and Engineering, Shanghai Jiao Tong University, China\\ $^2$Department of Computing Science, University of Aberdeen, UK\\ $^3$Institute of Cyber Security for Society (iCSS) \& School of Computing, University of Kent, UK\\
}

\begin{document}

\maketitle

\begin{abstract}
Distinguishing between human- and LLM-generated texts is crucial given the risks associated with misuse of LLMs. This paper investigates detection and explanation capabilities of current LLMs across two settings: binary (human vs. LLM-generated) and ternary classification (including an ``undecided'' class). We evaluate 6 close- and open-source LLMs of varying sizes and find that self-detection (LLMs identifying their own outputs) consistently outperforms cross-detection (identifying outputs from other LLMs), though both remain suboptimal. Introducing a ternary classification framework improves both detection accuracy and explanation quality across all models. Through comprehensive quantitative and qualitative analyses using our human-annotated dataset, we identify key explanation failures, primarily reliance on inaccurate features, hallucinations, and flawed reasoning. Our findings underscore the limitations of current LLMs in self-detection and self-explanation, highlighting the need for further research to address overfitting and enhance generalizability.
\end{abstract}

\section{Introduction}

The rise of large language models (LLMs) has brought remarkable advancements in natural language processing (NLP) tasks \cite{matarazzo2025survey}, including text generation. Models such as GPT-4o \cite{OpenAI2024GTP4O}, LLaMA \cite{Touvron2023LLaMA}, and Qwen \cite{ali2024qwen2} have blurred the boundaries between LLM-generated (LGTs) and human-generated texts (HGTs), posing new challenges in distinguishing between the two. While these capabilities of LLMs open new possibilities, they also bring concerns in areas such as misinformation, academic dishonesty, and automated content moderation \cite{hu2025unveiling}. As a result, detecting LGTs has become an increasingly important research area \cite{dugan2024raid, lee2023do, bhattacharjee2024fighting}.

Prior research has mainly focused on developing classifiers to distinguish HGTs and LGTs, including open-source detectors~\cite{Hans2024Binoculars} and online close-source detection systems~\cite{gptzero_website}. However, most detection systems have been limited to binary classification, which has several inherent issues. Recently, some works~\cite{lee2024llm} have attempted ternary classification by introducing a ``mixed'' category, which represents texts originating from mixed sources. However, this approach does not fundamentally resolve the issue. We further adopt the definition of an ``Undecided'' category based on other studies~\cite{ji2024detecting} and conduct ternary classification experiments for different LLMs, as certain texts are inherently indistinguishable between LGTs and HGTs. Furthermore, many studies treat the detection task as a black box, offering little insight into the decision-making process. Explainability, a critical aspect of trustworthy AI, has received less attention, but it is essential for building systems that users can trust \cite{weng2024understanding, zhou2024humanizing}. This paper presents an analysis of current LLMs in detecting LGTs and HGTs, with a particular emphasis on \textit{evaluating and improving the clarity of the explanations provided by LLM-based detectors}. By investigating how LLMs make predictions and offer explanations for their decisions, we aim to enhance their transparency and provide deeper insights into their reasoning processes.

This paper explores the explainability of LLM-based detectors, addressing two central questions: (1) \textit{How accurately can current LLM detectors identify origins of texts}, and (2) \textit{How reliable are their explanations?} Our study highlights that in a ternary setting compared to traditional binary classification, the average detection performance improves by 5.6\%, which demonstrates the necessity of ternary rather than binary setting to detect HGTs and LGTs. We further discovered that explanations are often flawed even when binary predictions are correct. Based on our comprehensive human-annotators' feedback, we summarize three common issues with explanations: reliance on inaccurate features (e.g., vague or irrelevant characteristics), hallucinations (e.g., non-existent or contradictory features), and incorrect reasoning (e.g., logical errors in attributing text origin). These explanation errors are quantified and categorized, with their distributions analyzed across different LLMs. Consequently, the proportion of explanation errors decreases by 13.3\% when we switch to ternary classification setting, which further supports the necessity of ternary classification for LGTs detection.

We evaluated 6 state-of-the-art (SOTA) LLM-based detectors, such as GPT-4o, GPT-4o mini, LLaMA3.3-70B, LLaMA3.3-7B, Qwen2-72B, and Qwen2-7B, on our created dataset comprising LGTs and HGTs. Moreover, our human annotators provided feedback based on correctness of predictions and explanations for this benchmark. Our results show that GPT-4o achieved the highest detection accuracy. In addition, LLMs performed better in self-detection than cross-detection, and ternary classification outperformed binary classification. Finally, explanation quality also improved under ternary setting, with fewer hallucinations and incorrect reasoning observed.

The main contributions of this work are:

\begin{itemize}[leftmargin=*, itemindent=0pt]
\item \textbf{Comprehensive Evaluation of Detection and Explanation:} We systematically assess current LLMs’ ability to detect and explain human- and LLM-generated texts using both binary and ternary classification tasks, demonstrating the advantages of ternary classification for both detection accuracy and explanation quality.
\item \textbf{Human-Annotated Dataset:} We present a new human-annotated dataset of LLM- and human-generated texts, enabling evaluation of LLM explanations and improved detector training.
\item \textbf{Analysis of Explanation Errors:} We quantitatively and qualitatively characterize key explanation failures, reliance on inaccurate features, hallucinations, and flawed reasoning, offering insights for LLM detection and self-explanation.
\end{itemize}


\begin{table*}[tb]
\centering
\small
\resizebox{\textwidth}{!}{
\begin{tabular}{lcccccc}
\toprule
\multirow{2}{*}{\diagbox{Datasets}{Models}} & \multicolumn{6}{c}{LLM Detectors}\\
& GPT-4o & GPT-4o mini & LLaMA3.3-70B & LLaMA3.3-8B & Qwen2-72B & Qwen2-7B \\
\midrule
\textbf{GPT-4o}      & \begin{tikzpicture}[baseline=(X.base)]
\node[fill=blue!30, inner sep=2pt, rounded corners] (X) {\textcolor{blue}{\textbf{71.39}}};\end{tikzpicture} & 59.38 & 57.31 & 48.41 & 64.14 & 59.76 \\ 
\textbf{GPT-4o mini}   & 65.71 & 61.03 & 53.75 & 51.73 & \textcolor{blue}{\textbf{67.27}} & 60.09 \\ 
\textbf{LLaMA3.3-70B}   & \textcolor{blue}{\textbf{67.26}} & 60.92 & \begin{tikzpicture}[baseline=(X.base)]
\node[fill=blue!30, inner sep=2pt, rounded corners] (X) {68.10};\end{tikzpicture} & 53.65 & 58.96 & 51.57 \\ 
\textbf{LLaMA3.3-8B}    & 60.74 & 55.77 & \textcolor{blue}{\textbf{62.29}} & \begin{tikzpicture}[baseline=(X.base)]
\node[fill=blue!30, inner sep=2pt, rounded corners] (X) {59.09};\end{tikzpicture} & 59.87 & 49.88 \\ 
\textbf{Qwen2-72B}      & 62.66 & \begin{tikzpicture}[baseline=(X.base)]
\node[fill=blue!30, inner sep=2pt, rounded corners] (X) {61.92};\end{tikzpicture} & 57.79 & 49.20 & \begin{tikzpicture}[baseline=(X.base)]
\node[fill=blue!30, inner sep=2pt, rounded corners] (X) {\textcolor{blue}{\textbf{68.15}}};\end{tikzpicture} & 61.36 \\ 
\textbf{Qwen2-7B}       & 62.45 & 59.06 & 59.12 & 48.57 & \textcolor{blue}{\textbf{65.24}} & \begin{tikzpicture}[baseline=(X.base)]
\node[fill=blue!30, inner sep=2pt, rounded corners] (X) {63.44};\end{tikzpicture} \\\midrule 
\textbf{Average}        & \textcolor{blue}{\textbf{65.03}} & 59.68 & 59.73 & 51.78 & 63.94 & 57.68 \\ 
\bottomrule
\end{tabular}}
\caption{F1 scores of LLM-based detectors in binary classification. The first column indicates different LLMs used for text generation, and the first row indicates different LLMs acting as detectors. The highest column-wise F1 score for each LLM detector to classify LGTs and HGTs across six datasets is highlighted in \colorbox{blue!30}{blue}. The highest row-wise F1 score for each LLM-generated text dataset across different LLM detectors is marked in \textcolor{blue}{\textbf{blue}}.}
\label{tab:binary_gt_performance}
\end{table*}
\section{Related Work}

\paragraph{LGT and HGT Detection.}
Past efforts to identify LGTs often relied on binary classification systems that distinguish HGTs from LGTs using surface-level features. While these methods were initially effective, they are prone to errors when encountering adversarial attacks or domain shifts, which limit their overall robustness~\cite{bhattacharjee2024fighting, dugan2024raid}. To address these limitations, researchers have explored strategies that integrate external knowledge, such as combining internal and external factual structures, to boost detection against diverse content and styles~\cite{ideate2024}. Recent studies also highlight the promise of using LLMs themselves for text detection: approaches like self-detection and mutual detection can outperform traditional classifiers, as illustrated by GPT-4's success in tasks like plagiarism detection~\cite{plagbench2024}. Notably, smaller models sometimes excel in zero-shot scenarios, offering adaptable solutions across varying architectures~\cite{smallerLLM2024}. Furthermore, \citet{lee2023do} demonstrated that LLMs can reliably identify their own outputs, providing a more nuanced framework for content verification. Despite these advances, the continuing challenges of domain adaptation and adversarial resistance underscore the need for more versatile and robust detection systems.

\textbf{Explainability in Detection Models.}
Recent work on LGT detection has focused on improving explainability. \citet{zhou2024humanizing} proposed to incorporate factual consistency into detection models to enhance their interpretability, while~\citet{weng2024understanding} explored mixed-initiative approaches that combine human expertise with automated models for better detection. These studies have made significant contributions to the field; however, they either depend heavily on expert input~\cite{weng2024understanding} or lack integration of explanation generation within the model itself~\cite{zhou2024humanizing}. Our approach, in contrast, enables LLMs to autonomously generate both predictions and detailed explanations, making it a more scalable and transparent solution for detecting machine-generated content.


\textbf{Ternary Classification.}
Traditional binary classification methods face limitations when texts exhibit ambiguous characteristics. Introducing an ``Undecided'' category addresses this by capturing three distinct scenarios (see Appendix~\ref{app:undecided_examples} for concrete illustrations): (1) \textit{Mixed texts} co-authored by humans and LLMs, where stylistic blending creates classification challenges; (2) \textit{Inherently ambiguous texts} that could plausibly originate from either source despite single authorship; and (3) \textit{Fragile indicators}, where subtle distinguishing features exist but lack robustness against behavioral evolution of either LLMs or human writers.
This approach advances beyond previous methods that primarily addressed mixed texts~\cite{lee2024llm}. The complexity is evidenced by Turing tests showing human difficulty in binary classification~\cite{turingtest2024}, and by studies demonstrating detector limitations with evolving writing patterns~\cite{coling2024}. The ternary framework improves both accuracy and explainability, particularly for these edge cases.

\section{LLM-based Binary Classification on LGTs and HGTs}
\label{sec:binary_classification}


\subsection{Experimental Design}

We selected six SOTA LLMs for text generation and subsequent detection: GPT-4o, GPT-4o mini~\cite{hurst2024gpt}, Qwen2-72B, Qwen2-7B~\cite{yang2024qwen2}, LLaMA3.3-70B, and LLaMA3.3-8B~\cite{dubey2024llama}. These LLMs were chosen for two main reasons. First, they represent the latest advancements in LLM development, demonstrating strong generation and detection capabilities. Second, selection spans different series and model sizes, enabling a comparative analysis of performance across architectures and scales.

To construct the dataset, we first selected 1,000 HGTs from publicly available M4GT-Bench dataset~\cite{wang2024m4gt}, ensuring a diverse range of topics, styles, and formats. Based on these selected HGTs, we designed 1,000 prompts that align with themes, structure, and style of the HGTs. Each LLM subsequently generated a corresponding response for each of these prompts. Together, these LGTs and HGTs formed the benchmark used in this study. For each text, the LLMs were tasked to determine its source (LGTs or HGTs) and provide an explanation, as illustrated in Table~\ref{tab:prompt_binary_classification}.

\begin{table}[h]
\centering
\small
\begin{tabular}{@{}p{\linewidth}@{}} 
\toprule
\textbf{Prompt}: Please determine whether the following text is generated by large language models or by a human, and provide a clear judgment. Additionally, please offer a detailed explanation for your decision. Please structure your answer in JSON format as follows: \{``answer'': , ``explanation'': \}.\\
\bottomrule
\end{tabular}
\caption{A prompt for LLMs to determine text origin and provide an explanation under a binary setting.}
\label{tab:prompt_binary_classification}
\end{table}
\begin{table*}[tb]
\centering
\small
\resizebox{\textwidth}{!}{
\begin{tabular}{lcccccc}
\toprule
\multirow{2}{*}{\diagbox{Datasets}{Models}} & \multicolumn{6}{c}{LLM Detectors}\\
& GPT-4o & GPT-4o mini & LLaMA3.3-70B & LLaMA3.3-8B & Qwen2-72B & Qwen2-7B \\
\midrule
\textbf{GPT-4o}      & \begin{tikzpicture}[baseline=(X.base)]
\node[fill=blue!30, inner sep=2pt, rounded corners] (X) {\textcolor{blue}{\textbf{73.48}}};\end{tikzpicture}
/ \begin{tikzpicture}[baseline=(X.base)]
\node[fill=red!30, inner sep=2pt, rounded corners] (X) {\textcolor{red}{\textbf{67.04}}};
\end{tikzpicture} & 58.41 / 54.17 & 57.65 / 54.17 & 48.29 / 51.32 & 64.11 / 59.60 & 59.57 / \begin{tikzpicture}[baseline=(X.base)]
\node[fill=red!30, inner sep=2pt, rounded corners] (X) {62.13};\end{tikzpicture} \\ 
\textbf{GPT-4o mini}   & 63.72 / 60.95 & \begin{tikzpicture}[baseline=(X.base)]
\node[fill=blue!30, inner sep=2pt, rounded corners] (X) {63.43};\end{tikzpicture} / 60.15 & 53.85 / 52.46 & 50.01 / 47.75 & \textcolor{blue}{\textbf{66.91}} / \textcolor{red}{\textbf{61.08}} & 60.13 / 57.28 \\ 
\textbf{LLaMA3.3-70B}   & 68.13 / 63.96 & 62.12 / 60.32 & \begin{tikzpicture}[baseline=(X.base)]
\node[fill=blue!30, inner sep=2pt, rounded corners] (X) {\textcolor{blue}{\textbf{68.33}}};\end{tikzpicture} / \begin{tikzpicture}[baseline=(X.base)]
\node[fill=red!30, inner sep=2pt, rounded corners] (X) {\textcolor{red}{\textbf{64.47}}};\end{tikzpicture} & 53.13 / 51.78 & 58.16 / 59.22 & 52.41 / 48.82 \\ 
\textbf{LLaMA3.3-8B}    & 58.97 / 61.11 & 56.19 / 55.98 & \textcolor{blue}{\textbf{63.24}} / \textcolor{red}{\textbf{63.72}} & \begin{tikzpicture}[baseline=(X.base)]
\node[fill=blue!30, inner sep=2pt, rounded corners] (X) {58.73};\end{tikzpicture} / \begin{tikzpicture}[baseline=(X.base)]
\node[fill=red!30, inner sep=2pt, rounded corners] (X) {56.29};\end{tikzpicture} & 59.83 / 59.17 & 49.66 / 48.97 \\ 
\textbf{Qwen2-72B}      & 62.70 / 61.09 & 62.91 / \begin{tikzpicture}[baseline=(X.base)]
\node[fill=red!30, inner sep=2pt, rounded corners] (X) {62.84};\end{tikzpicture} & 58.71 / 56.99 & 49.12 / 47.26 & \begin{tikzpicture}[baseline=(X.base)]
\node[fill=blue!30, inner sep=2pt, rounded corners] (X) {\textcolor{blue}{\textbf{70.47}}};\end{tikzpicture} / \begin{tikzpicture}[baseline=(X.base)]
\node[fill=red!30, inner sep=2pt, rounded corners] (X) {\textcolor{red}{\textbf{67.98}}};\end{tikzpicture} & 61.24 / 58.18 \\ 
\textbf{Qwen2-7B}       & 63.78 / 61.54 & 58.11 / 57.84 & 60.15 / 58.60 & 48.72 / 49.17 & \textcolor{blue}{\textbf{65.44}} / \textcolor{red}{\textbf{63.58}} & \begin{tikzpicture}[baseline=(X.base)]
\node[fill=blue!30, inner sep=2pt, rounded corners] (X) {63.83};\end{tikzpicture} / 61.91 \\ 
\bottomrule
\end{tabular}}
\caption{F1 scores of LLM-based detectors on human-annotated texts for \textbf{binary} classification. Each dataset contains 100 LGTs and 100 HGTs with human-annotated explanations. Each cell indicates classification/explanation F1, where the highest column-wise F1 of each LLM detector for binary classification and explanations across different generated texts are highlighted with \colorbox{blue!30}{blue} and \colorbox{red!30}{red}, respectively. In addition, the highest row-wise F1 among different LLM detectors for each LLM-generated text datasets are indicated with \textcolor{blue}{\textbf{blue}} and \textcolor{red}{\textbf{red}} in bold, respectively.}
\label{tab:binary_annotate_performance}
\end{table*}
\paragraph{Manual Annotation.} To assess LLMs' ability to explain text origins and identify distinguishing features, 3 co-authors, who are undergraduate computer science students, manually evaluated correctness of LLM-generated explanations. They determined accuracy of each explanation. From 7 datasets (6 SOTA LLMs + Human), 100 texts with corresponding explanations per dataset were randomly selected for human evaluation. All annotators assessed explanations provided by each model, which achieved a Fleiss' kappa~\cite{Fleiss1971kappa} of 0.8387, indicating near-complete agreement. Annotation guidelines are detailed in Appendix~\ref{appendix:annotation_guidelines}.

\paragraph{Evaluation Metrics.} For evaluating the classification performance of the LLMs, the primary metric we used is the F1 score. To assess the quality of explanations, human evaluators reviewed the LLM-generated explanations and classified them as correct or incorrect. The F1 score was also used as the evaluation metric for explanation quality.

\subsection{Binary Classification Results}

We evaluated the performance of six LLMs across six datasets, as detailed in Table~\ref{tab:binary_gt_performance}, which systematically compares the detection capabilities of various LLMs for both LGTs and HGTs. The results demonstrate that GPT-4o achieves the best average detection performance across all datasets, showing relatively strong generalization capabilities. Larger parameter models generally exhibit significantly better detection performance than smaller ones, which suggests that these models are not merely making random guesses but are effectively identifying distinctive textual features.

The \colorbox{blue!30}{F1 scores} in Table~\ref{tab:binary_gt_performance}'s diagonal direction show that LLMs within same series consistently detect their own outputs more effectively than those from other LLM families. For example, LLaMA3.3 70B achieves the highest \colorbox{blue!30}{F1 score} in its generated dataset, which indicates a heightened sensitivity to its own text distribution compared to other LLMs. However, this specialization reduces cross-detection performance, as seen in Qwen2-7B's lower F1 on LLaMA-generated texts. While larger LLMs generally achieve better detection across different LLMs, such as GPT-4o, GPT-4o mini, LLaMA3.3-70B and Qwen2-72B, their outputs are also more difficulty to distinguish by smaller LLMs, such as LLaMA3.3-8B and Qwen2-7B.

Additionally, based on the human annotations of sampled 100 LGTs and 100 HGTs with explanations from each dataset, we observed that the detection and explanation results across different LLMs are not entirely consistent, as shown in Table~\ref{tab:binary_annotate_performance}. We noted that in some cases, the F1 score for explanations was higher than that for classification. This is because, in these cases, the explanation correctly identified the reasoning for attribution, but the final classification was incorrect. For instance, the difference in F1 scores between explanation and classification was particularly noticeable for LLaMA3.3-8B and Qwen2-7B, suggesting that these models struggle to truly comprehend the textual features necessary for correctly determining the origin of generated texts, which results in lower detection performance.

As shown in Table~\ref{tab:binary_mgt_hgt_performance}, analysis of the annotators' results revealed that models are generally more accurate in attributing HGTs compared to LGTs. For example, while GPT-4o demonstrates higher accuracy (78 out of 100) in classifying HGTs, the false explanations account for more than 47\%.

\begin{table*}[tb]
\centering
\footnotesize
\renewcommand{\arraystretch}{1.2} 
\resizebox{\textwidth}{!}{
\begin{tabular}{lcccccccccccc}
\toprule
\multirow{2}{*}{Model} & \multicolumn{6}{c}{MGTs} & \multicolumn{6}{c}{HGTs} \\
\cmidrule(lr){2-4} \cmidrule(lr){5-7} \cmidrule(lr){8-10} \cmidrule(lr){11-13}
 & TC & TE & FE & FC & TE & FE & TC & TE & FE & FC & TE & FE \\
\midrule
GPT-4o & 64 & ${51}_{\textcolor{teal}{:79.7\%}}$ & ${13}_{\textcolor{teal}{:20.3\%}}$ & 36 & ${8}_{\textcolor{teal}{:22.2\%}}$ & ${28}_{\textcolor{teal}{:77.8\%}}$ & 78 & ${41}_{\textcolor{teal}{:52.6\%}}$ & ${37}_{\textcolor{teal}{:47.4\%}}$ & 22 & ${2}_{\textcolor{teal}{:9.1\%}}$ & ${20}_{\textcolor{teal}{:90.9\%}}$ \\
LLaMA3.3-70B & 56 & ${35}_{\textcolor{teal}{:62.5\%}}$ & ${21}_{\textcolor{teal}{:37.5\%}}$ & 44 & ${10}_{\textcolor{teal}{:22.7\%}}$ & ${34}_{\textcolor{teal}{:77.3\%}}$ & 60 & ${37}_{\textcolor{teal}{:61.7\%}}$ & ${23}_{\textcolor{teal}{:38.3\%}}$ & 40 & ${7}_{\textcolor{teal}{:17.5\%}}$ & ${33}_{\textcolor{teal}{:82.5\%}}$ \\
Qwen2-72B & 60 & ${36}_{\textcolor{teal}{:60.0\%}}$ & ${24}_{\textcolor{teal}{:40.0\%}}$ & 40 & ${7}_{\textcolor{teal}{:17.5\%}}$ & ${33}_{\textcolor{teal}{:82.5\%}}$ & 69 & ${50}_{\textcolor{teal}{:72.5\%}}$ & ${19}_{\textcolor{teal}{:27.5\%}}$ & 31 & ${7}_{\textcolor{teal}{:22.6\%}}$ & ${24}_{\textcolor{teal}{:77.4\%}}$ \\
\bottomrule
\end{tabular}}
\caption{Performance of LLMs on LLM-generated and human-generated texts for ternary classification and explanation tasks. It includes results for classification and explanation tasks, where TC represents true classification, FC represents false classification, TE represents true explanation, and FE represents false explanation. Note: TC=TE+FE and FC=TE+FE.}
\label{tab:binary_mgt_hgt_performance}
\end{table*}
\begin{table*}[tb]
\centering
\small
\resizebox{\textwidth}{!}{
\begin{tabular}{lcccccc}
\toprule
\multirow{2}{*}{\diagbox{Datasets}{Models}} & \multicolumn{6}{c}{LLM Detectors}\\
& GPT-4o & GPT-4o mini & LLaMA3.3-70B & LLaMA3.3-8B & Qwen2-72B & Qwen2-7B \\
\midrule
\textbf{GPT-4o} & \begin{tikzpicture}[baseline=(X.base)]
\node[fill=blue!30, inner sep=2pt, rounded corners] (X) {\textcolor{blue}{\textbf{79.73}}};\end{tikzpicture}/\textcolor{red}{\textbf{72.04}} & 64.62/61.87 & 62.19/59.04 & 58.06/57.78 & 71.62/68.86 & 63.81/62.72 \\
\textbf{GPT-4o mini} & \textcolor{blue}{\textbf{70.11}}/\textcolor{red}{\textbf{68.75}} & \begin{tikzpicture}[baseline=(X.base)]
\node[fill=blue!30, inner sep=2pt, rounded corners] (X) {67.39};\end{tikzpicture}/  \begin{tikzpicture}[baseline=(X.base)]
\node[fill=red!30, inner sep=2pt, rounded corners] (X) {65.18};\end{tikzpicture} & 58.88/52.95 & 54.43/51.16 & 69.65/65.95 & 65.15/62.60 \\
\textbf{LLaMA3.3-70B} & \textcolor{blue}{\textbf{74.41}}/\begin{tikzpicture}[baseline=(X.base)]
\node[fill=red!30, inner sep=2pt, rounded corners] (X) {\textcolor{red}{\textbf{75.26}}};\end{tikzpicture} & 65.16/64.75 & \begin{tikzpicture}[baseline=(X.base)]
\node[fill=blue!30, inner sep=2pt, rounded corners] (X) {72.11};\end{tikzpicture}/ \begin{tikzpicture}[baseline=(X.base)]
\node[fill=red!30, inner sep=2pt, rounded corners] (X) {71.83};\end{tikzpicture} & 57.05/57.34 & 64.94/62.44 & 56.46/55.32 \\
\textbf{LLaMA3.3-8B} & \textcolor{blue}{\textbf{71.99}}/\textcolor{red}{\textbf{70.80}} & 60.18/61.10 & 64.82/63.93 & \begin{tikzpicture}[baseline=(X.base)]
\node[fill=blue!30, inner sep=2pt, rounded corners] (X) {63.96};\end{tikzpicture}/ \begin{tikzpicture}[baseline=(X.base)]
\node[fill=red!30, inner sep=2pt, rounded corners] (X) {62.85};\end{tikzpicture} & 63.12/60.52 & 54.08/53.01 \\
\textbf{Qwen2-72B} & 67.28/66.74 & 65.12/64.73 & 61.81/61.74 & 53.24/52.87 & \begin{tikzpicture}[baseline=(X.base)]
\node[fill=blue!30, inner sep=2pt, rounded corners] (X) {\textcolor{blue}{\textbf{76.05}}};\end{tikzpicture}/ \begin{tikzpicture}[baseline=(X.base)]
\node[fill=red!30, inner sep=2pt, rounded corners] (X) {\textcolor{red}{\textbf{75.56}}};\end{tikzpicture} & 65.26/64.72 \\
\textbf{Qwen2-7B} & 68.91/67.42 & 60.15/59.31 & 62.06/61.57 & 52.41/52.30 & \textcolor{blue}{\textbf{70.30}}/\textcolor{red}{\textbf{68.44}} & \begin{tikzpicture}[baseline=(X.base)]
\node[fill=blue!30, inner sep=2pt, rounded corners] (X) {\textcolor{blue}{\textbf{66.61}}};\end{tikzpicture}/ \begin{tikzpicture}[baseline=(X.base)]
\node[fill=red!30, inner sep=2pt, rounded corners] (X) {65.17};\end{tikzpicture} \\
\bottomrule
\end{tabular}}
\caption{F1 scores of LLM-based detectors on the \textbf{ternary} classification of LGTs and HGTs. The highest column-wise and row-wise F1 scores are highlighted and marked following the same scheme as in Table \ref{tab:binary_annotate_performance}.}
\label{tab:ternary_performance}
\end{table*}
\begin{table}[tb]
\centering
\small
\begin{tabular}{@{}p{\linewidth}@{}} 
\toprule
\textbf{Prompt}: Please classify the following text into one of three categories based on its source: LLM-generated, human-generated, or undecided. The ``Undecided'' category refers to texts that exhibit characteristics of both LLM-generated and human-generated content, making it impossible and inappropriate to distinguish between the two. Provide a clear classification and a detailed explanation for your decision. Structure your answer in JSON format as follows: \{``classification'': , ``explanation'': \}.\\
\bottomrule
\end{tabular}
\caption{A prompt for LLMs to determine text origin and provide an explanation under a ternary setting.}
\label{tab:prompt_ternary_classification}
\end{table} 
\section{LLM-based Ternary Classification on LGTs and HGTs}

\subsection{Experimental Setup}

Using the same benchmark in~\S~\ref{sec:binary_classification}, we prompted the LLMs for ternary classification and the prompt template is demonstrated in Table~\ref{tab:prompt_ternary_classification}. The ground truth for the ternary classification was determined based on annotators' votes, where the three annotators were aware of the text's origin (LLMs or human) and were asked to distinguish between the ground truth and the ``Undecided'' category. This allowed for the evaluation of both the LLM's classification results and the explanations provided by the LLMs. The Fleiss' kappa~\cite{Fleiss1971kappa} for the ternary classification annotations among the three annotators was calculated as 0.7629, which indicates substantial agreement.

\subsection{Ternary Classification Results}

Table~\ref{tab:ternary_performance} presents the F1 scores of LLMs in the ternary classification setting. Comparing it with Table~\ref{tab:binary_annotate_performance}, we observe that introducing the ``Undecided'' category leads to overall performance improvements across both classification and explanation tasks. Specifically, GPT-4o exhibits the most notable gains, improving from 73.48/67.04 to 79.73/72.04, indicating that a finer-grained classification allows stronger models to better capture nuanced differences between LGTs and HGTs.

Moreover, Fig~\ref{fig:ternary_confusion_matrices} reveals how different models distribute predictions across three categories. GPT-4o demonstrates a more balanced distribution, with relatively lower misclassification rates for both HGTs and LGTs. In contrast, LLaMA3-70B shows a stronger tendency to label texts as ``human-generated'', leading to a higher false positive rate. Meanwhile, Qwen2-72B exhibits a more cautious classification approach, assigning a larger proportion of texts to ``Undecided'' category, particularly for LGTs.

\pgfplotsset{
    CM common axis styles/.style={
        width=1.8in,
        height=1.8in,
        xmajorgrids=false,
        ymajorgrids=false,
        xmin=1, xmax=3,
        ymin=1, ymax=3,
        xtick={1,2,3},
        ytick={1,2,3},
        xticklabels={LLMs, Undecided, Human},
        yticklabels={LLMs, Undecided, Human},
        x tick label style={font=\scriptsize},
        y tick label style={font=\scriptsize},
        tickwidth=0pt,
        tick align=outside,
        enlargelimits={abs=0.5},
        colormap={whiteblue}{rgb255(0cm)=(255,255,255) rgb255(1cm)=(128,128,255)}
        },
    CM common plot styles/.style={
        matrix plot,
        mesh/cols=3,
        mesh/rows=3,
        point meta=explicit,
        nodes near coords,
        nodes near coords align={center},
        visualization depends on={value \thisrow{q} \as \rawvalue},
        nodes near coords style={font=\tiny, color=black}
        }
}
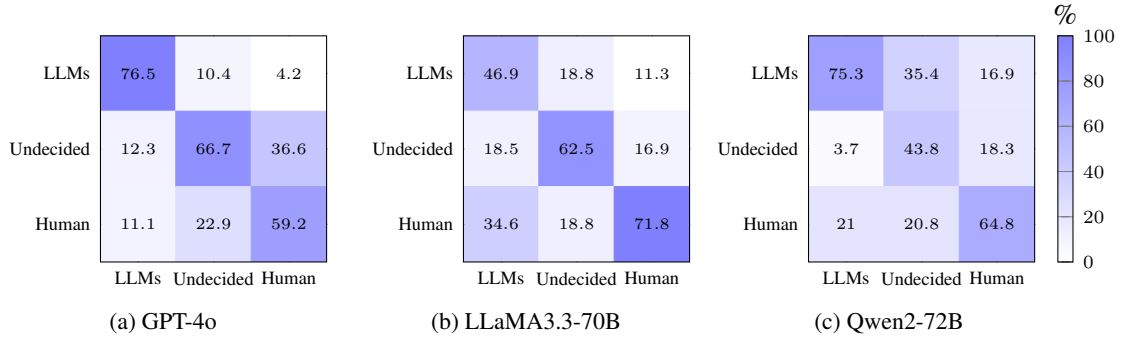
\begin{figure*}[tb]
\begin{subfigure}{0.3\textwidth} 
    \centering
    \begin{tikzpicture}
    \begin{axis}[
    CM common axis styles
    ]
    \addplot [
    CM common plot styles
    ] table [meta=p] {
    x y p
    1 1 76.5
    2 1 10.4
    3 1 4.2
    1 2 12.3
    2 2 66.7
    3 2 36.6
    1 3 11.1
    2 3 22.9
    3 3 59.2
    };
    \end{axis}
    \end{tikzpicture}
    \caption{GPT-4o}
\end{subfigure}%
\begin{subfigure}{0.3\textwidth} 
    \centering
    \begin{tikzpicture}
    \begin{axis}[
    CM common axis styles
    ]
    \addplot [
    CM common plot styles
    ] table [meta=p] {
    x y p
    1 1 46.9
    2 1 18.8
    3 1 11.3
    1 2 18.5
    2 2 62.5
    3 2 16.9
    1 3 34.6
    2 3 18.8
    3 3 71.8
    };
    \end{axis}
    \end{tikzpicture}
    \caption{LLaMA3.3-70B}
\end{subfigure}%
\begin{subfigure}{0.3\textwidth} 
    \centering
    \begin{tikzpicture}
    \begin{axis}[
    CM common axis styles,
    colorbar, 
    colorbar style={
        width=2mm, 
        ytick={0, 20, 40, 60, 80, 100},
        yticklabel style={font=\scriptsize},
        title=\%,
        title style={
            yshift=-2mm  
        }
    },
    point meta min=0,
    point meta max=100
    ]
    \addplot [
    CM common plot styles
    ] table [meta=p] {
    x y p
    1 1 75.3
    2 1 35.4
    3 1 16.9
    1 2 3.7
    2 2 43.8
    3 2 18.3
    1 3 21.0
    2 3 20.8
    3 3 64.8
    };
    \end{axis}
    \end{tikzpicture}
    \caption{Qwen2-72B}
\end{subfigure}
\caption{Confusion matrices showing the performance of LLM-based detectors (GPT-4, LLaMA3.3-70B, Qwen2-72B) in the ternary classification task, where the bottom row represents human-annotated ground truth labels (LGTs, Undecided, and HGTs), and the left column represents classification results predicted by LLM-based detectors.}
\label{fig:ternary_confusion_matrices}
\end{figure*}
\begin{table*}[tb]
\centering
\small
\begin{tabular}{@{}p{\linewidth}@{}}
\toprule
\textbf{Example 1: Ambiguous Features}

\textbf{Original Text}: ``Quantum entanglement enables faster-than-light communication, allowing for instantaneous data transfer across vast distances. This phenomenon has vast implications for future communication technologies, making it a subject of growing interest in both academic and technological circles.''

\textbf{LLM Judgment}: LLM-generated

\textbf{LLM Explanation}: ``The text discusses highly technical content, filled with specialized scientific terms, which is characteristic of machine-generated text. Such advanced topics are often produced by AI systems rather than human experts.''\\

\midrule

\textbf{Example 4: Incorrectly Perceived Repetition}

\textbf{Original Text}: ``The discovery of gravitational waves was a turning point in the history of physics. This breakthrough confirmed Einstein’s theory of relativity, providing compelling evidence for the existence of gravitational waves, which opened new doors in astrophysics research.''

\textbf{LLM Judgment}: LLM-generated

\textbf{LLM Explanation}: ``The text exhibits repetitive sentence structures, a hallmark of AI-generated content. The repeated mention of 'gravitational waves' and 'breakthrough' indicates a pattern commonly seen in machine-generated texts.''\\

\midrule

\textbf{Example 6: Misinterpretation of Formal Tone}

\textbf{Original Text}: ``One must consider the economic ramifications of such policies, as they extend beyond immediate financial concerns into broader societal impacts. The implications of these policies are far-reaching, touching on employment, income inequality, and long-term societal well-being.''

\textbf{LLM Judgment}: LLM-generated

\textbf{LLM Explanation}: ``The formal tone and structured language initially suggest human authorship, as such features are often attributed to human experts. However, LLMs can replicate this style with high fidelity, leading to the final classification as LLM-generated.''\\

\bottomrule
\end{tabular}
\caption{Analysis of LLM vs. Human Writing Attribution Based on Various Features. The table categorizes examples where LLM-generated and human-written texts were incorrectly attributed or analyzed, providing explanations and analyses of these misattributions.}
\vspace{-1.5em}
\label{tab:main_misattributions_examples}
\end{table*}
A closer comparison between binary and ternary classifications in Tables~\ref{tab:binary_annotate_performance} and \ref{tab:ternary_performance} suggests that added ``Undecided'' category benefits models differently. While large models like GPT-4o and LLaMA3-70B leverage this additional flexibility to improve both classification and explanation F1 scores, smaller models such as Qwen2-7B show more mixed results, with only marginal improvements. This suggests that high-capacity models may be better equipped to handle ambiguous cases, while smaller models struggle with added complexity.

Overall, these findings indicate that ternary classification not only refines detection performance but also enhances the LLMs' ability to generate more meaningful explanations. The improvements are particularly evident in large-scale LLMs, which benefit from a more nuanced decision space.

\section{Explainability of LLM-based Detectors}

\subsection{Incorrect Explanation Attribution}

Although LLMs can distinguish LGTs and HGTs, especially in self-detection settings, there are explanations that are incorrect via human evaluation. Normally, incorrect explanations in correctly classified cases fall into three types: inaccurate features (misidentifying key attributes), hallucinations (citing nonexistent or contradictory features), and flawed reasoning (faulty logic despite a correct outcome). For misclassified texts, errors typically involve inaccurate features or hallucinations, which highlights the need to prioritize explanation accuracy alongside detection performance to enhance trust in LLM-based detectors.
\begin{table*}[tb]
\centering
\small
\resizebox{\textwidth}{!}{
\begin{tabular}{lcccccc}
\toprule
Models & Ambiguous Features (\%) & Surface Features (\%) & Logic\&Emotion (\%) & Vocabulary (\%) & Hallucinations (\%) & Incorrect Reasoning (\%)\\
\midrule
GPT-4o & 32.7 & 12.4 & 43.2 & 7.2 & 2.2 & 2.3\\
LLaMA3.3-70B & 40.1 & 25.7 & 18.9 & 4.1 & 6.1 & 5.1\\
Qwen2-72B & 32.1 & 23.7 & 25.4 & 8.9 & 6.3 & 3.6\\
\bottomrule
\end{tabular}}
\caption{Attribution differences among LLMs when the judgment is correct but the explanation is incorrect.}
\label{tab:correct_but_wrong_explanation}
\end{table*}

\begin{table*}[tb]
\centering
\small
\resizebox{\textwidth}{!}{
\begin{tabular}{lccccc}
\toprule
Models & Ambiguous Features (\%) & Surface Features (\%) & Logic\&Emotion (\%) & Vocabulary (\%) & Hallucinations (\%)\\
\midrule
GPT-4o & 13.9 & 30.7 & 23.8 & 20.7 & 10.9\\
LLaMA3.3-70B & 26.8 & 10.1 & 40.1 & 4.1 & 18.9\\
Qwen2-72B & 33.7 & 20.1 & 9.9 & 26.4 & 9.9\\
\bottomrule
\end{tabular}}
\caption{Attribution differences among LLMs when both the judgment and the explanation are incorrect.}
\vspace{-1.5em}
\label{tab:incorrect_and_wrong_explanation}
\end{table*}
\paragraph{Inaccurate Features.}
Incorrect explanation attribution is often caused by relying on ambiguous, superficial, or misinterpreted features, as shown in the examples in Tables~\ref{tab:main_misattributions_examples} and \ref{tab:misattributions_examples} of Appendix~\ref{appendix:misattributions_examples}.
In Example 1 ``Ambiguous Features'', the model misclassifies text on quantum entanglement as LGTs due to technical jargon usage. However, advanced topics can also be written by human experts, not just LLMs. This text was actually human-generated. Similarly, Example 2 ``Surface Features'' shows how the model links grammatical errors to the machine. Such mistakes are common among both native and non-native writers and should not be sole indicators of LGTs. In fact, HGTs are more likely to contain grammatical errors.
Example 3 illustrates a misjudgment where emotional complexity is falsely attributed exclusively to human writing. The model assumes nuanced emotional contrasts inherently reflect human authorship, overlooking modern LLMs' capability to simulate such depth. This case underscores the unreliability of using emotional sophistication alone as a criterion to differentiate between HGTs and LGTs.

\paragraph{Hallucinations.}
Hallucinations occur when the model incorrectly attributes features to the text that either do not exist or are contrary to the actual content. In Example 4: Incorrectly Perceived Repetition, the model misinterprets the repetition of ideas about gravitational waves as a sign of LLM authorship. The text does not exhibit excessive repetition, and the claim of a repetitive structure is a false attribution, likely due to biases in the model’s training data.
In Example 5: Fictitious Absence of Domain Knowledge, the model mistakenly claims that a text about RNA interference lacks technical depth, suggesting it is more likely to be human-written. In reality, the text contains domain-specific biological content, and the model fails to recognize the technical knowledge present.

\paragraph{Incorrect Reasoning.}
Incorrect reasoning occurs when relevant features are correctly identified but are misinterpreted, leading to incorrect conclusions. Example 6 highlights a classification error rooted in inconsistent reasoning. The model correctly identifies formal stylistic features but misapplies their significance. Enforcing a binary classification may lead to inconsistent reasoning in model’s inference process, as it forces an erroneous LLM label despite ambiguity that could be better captured in a ternary framework.

\subsection{Human Evaluation}

The reasons for incorrect explanations from human annotators are categorized into two scenarios: correct predictions with incorrect explanations and incorrect predictions with incorrect explanations. The results are summarized in Tables~\ref{tab:correct_but_wrong_explanation} and \ref{tab:incorrect_and_wrong_explanation}.

For cases where the model made correct predictions but provided incorrect explanations, Table~\ref{tab:correct_but_wrong_explanation} shows that the most prevalent reasons were inaccurate features and hallucinations. Inaccurate features, such as attributing the decision to vague or irrelevant characteristics, accounted for a significant portion of errors across all LLMs. Hallucinations were also frequent, particularly for models like Qwen2-7B and GPT-4o. Faulty reasoning, though less common, contributed to the proportion of incorrect explanations, highlighting inconsistencies in reasoning despite identifying correct features.
For cases involving both incorrect predictions and incorrect explanations, Table~\ref{tab:incorrect_and_wrong_explanation} indicates a similar distribution of error types, but with a higher prevalence of hallucinations. Annotators noted that models hallucinated key features, attributing the decision to features not present in the text, which compounded the issue of misclassification. 

Overall, the analysis reveals that hallucinations and reliance on inaccurate features are dominant sources of error in explanations, regardless of prediction accuracy. Addressing these issues requires further refinement of the interpretability mechanisms in LLMs, with a focus on grounding explanations in verifiable and relevant textual evidence.

\begin{table*}[tb]
\centering
\small
\resizebox{\textwidth}{!}{
\begin{tabular}{lcccccc}
\toprule
\multirow{2}{*}{Model} & \multicolumn{6}{c}{Datasets} \\
& GPT-4o & GPT-4o mini & LLaMA3.3-70B & LLaMA3.3-8B & Qwen2-72B & Qwen2-7B \\
\midrule
\textbf{Qwen2-7B (base model)} 
& 59.57 & 60.13 & 52.41 & 49.66 & 61.24 & 63.83 \\

\textbf{SFT (only answer)} 
& & & & & & \\
\quad w/ LLaMA3.3-8B 
& 58.28 & 60.17 & 52.57 & 49.85 & 55.41 & 59.52 \\

\quad w/ GPT-4o 
& 63.70 & 62.18 & 53.90 & 52.91 & 63.28 & 62.86 \\

\quad w/ mixed dataset (ALL) 
& 62.87 & 63.10 & 53.28 & 54.33 & 64.17 & 67.91 \\

\textbf{SFT (answer \& explanation)} 
& & & & & & \\
\quad w/ GPT-4o 
& \underline{70.09} & \textbf{72.03} & 58.19 & 54.17 & 67.39 & 68.55 \\

\quad w/ mixed dataset (ALL) 
& 68.14 & \underline{71.09} & \underline{60.41} & 54.59 & \underline{70.72} & \underline{70.90} \\

\textbf{GRPO} 
& & & & & & \\
\quad w/o cold-start 
& 63.31 & 64.17 & 57.96 & \underline{56.09} & 63.60 & 63.02 \\

\quad w/ cold-start 
& \textbf{73.29} & 69.01 & \textbf{63.34} & \textbf{65.22} & \textbf{74.89} & \textbf{73.47} \\
\bottomrule
\end{tabular}}
\caption{
F1 score comparison of various training strategies across different evaluation models. 
\textbf{Qwen2-7B (base model)} represents the raw performance without fine-tuning. 
\textbf{SFT (only answer)} and \textbf{SFT (answer \& explanation)} denote models supervised-fine-tuned using answer-only or answer-plus-explanation data, respectively. 
In ``SFT w/ GPT-4o'' or ``w/ LLaMA3.3-8B'', the training dataset was generated using the outputs of the specified LLM. 
``Mixed dataset (ALL)'' combines training data from multiple sources. 
\textbf{GRPO} refers to the Group Relative Policy Optimization, evaluated with and without cold-start initialization. 
\textbf{Bold} indicates the best score per column; \underline{underline} indicates the second-best.
}
\vspace{-1.5mm}
\label{tab:grpo_ablation}
\end{table*}
\section{Can we improve LLM-based LGT detection and explanation?}

\paragraph{Supervised Fine-Tuning and Reinforcement Learning.}
To investigate effectiveness of supervised fine-tuning (SFT) on improving cross-LLM detection, we conducted a series of experiments using datasets generated from different LLMs, with or without explanations. Details of data construction are provided in Appendix~\ref{app:sft}, and results are summarized in Table~\ref{tab:grpo_ablation}. When using answer-only data for SFT, we observe a clear advantage when training data is generated by GPT-4o compared to LLaMA3.3-8B. This likely reflects higher quality and linguistic diversity of GPT-4o generations, whereas LLaMA3.3-8B tends to produce simpler and less informative outputs, offering limited supervision signals. Furthermore, although models fine-tuned on GPT-4o-only data generally achieve higher F1 scores when evaluated on GPT-family outputs, the mixed-source dataset provides stronger generalization across all target LLMs. This indicates that exposure to varied generation styles during training helps the model better capture cross-LLM decision boundaries. Additionally, incorporating explanations into training data consistently yields higher performance. The improvement suggests that explanation-augmented SFT encourages model to internalize task-relevant reasoning patterns and enhances its ability to focus on discriminative linguistic cues indicative of text origin, rather than merely memorizing surface features.

Motivated by recent advances in reward-optimized reasoning, such as OpenAI-o1\cite{openai2024o1} and DeepSeek-R1\cite{deepseek2025r1}, we further explore the impact of RL on cross-LLM detection. Specifically, we adopt a GRPO-style reward optimization framework inspired by DeepSeek-R1, with implementation details in Appendix~\ref{app:rl}. As shown in Table~\ref{tab:grpo_ablation}, applying GRPO without any SFT initialization already brings consistent gains over base model across all test LLMs, with the most notable improvement observed on LLaMA3.3-8B. However, performance remains generally lower than the best SFT configurations. When we initialize GRPO with a model that has been fine-tuned using explanation-augmented, mixed-source data (i.e., the cold-start setting), we observe substantial performance boosts across all evaluation datasets. This combination effectively leverages reasoning capacity learned during SFT and further refines it through reward-driven preference learning. Results demonstrate that GRPO with a well-informed initialization can significantly enhance detection accuracy, enabling the model to better align its scoring behavior with human-intuitive criteria for text provenance.

\paragraph{Enhancing LLM Detection through LLM Collaboration.}
We further explore whether the performance of LLM-based detection can improve via LLM's collaboration. Table~\ref{tab:model_collaboration} shows that the performance of LLM-based detectors improves significantly when their judgments and explanations are complemented by another LLM counterpart. Specifically, the cross-detection of GPT-4o on Qwen-2 72B dataset has noticeable improvements in the classification and explanation F1 with the support of Qwen2-72B. We also find a similar trend, where Qwen2-72B benefits from GPT-4o's support on the cross-detection settings. These findings indicate that LLM's collaboration can further improve the classification and explanation performance on the LLM  counterpart's dataset, i.e., cross-detection.


\begin{table}[ht] 
\small 
\centering 
\resizebox{\columnwidth}{!}{
 \begin{tabular}{lcc}
\toprule 
\diagbox{Models}{Datasets} & GPT-4o & Qwen2-72B\\
\midrule 
GPT-4o (+Qwen2-72B) & +1.43\% / -0.61\% & +3.79\% / +2.33\%\\
Qwen2-72B (+GPT-4o) & +2.01\% / +0.45\% & +0.39\% / -1.24\%\\
\bottomrule 
\end{tabular} 
}
\caption{F1 score differences based on LLM collaboration with judgments and explanations integration. The supplemental LLMs in () will generate the judgments and explanations first to further support the detection and explanations of main LLMs.} 
\label{tab:model_collaboration} 
\end{table}





\section{Conclusion}
We evaluated how well LLM-based detectors differentiate human- from LLM-generated text, focusing on detection accuracy and explanation clarity. Self-detection by LLM-based detectors reliably outperforms cross-detection, especially within the same model family. Yet their explanations remain flawed, hinging on spurious features, hallucinations, and unsound reasoning, with GPT-4o trading higher accuracy for frequent hallucinations and Qwen2-7B offering a more balanced but vague rationale. These results underscore the imperative for more interpretable, trustworthy detectors in critical applications such as academic integrity and content moderation.

\section*{Limitations}

This study is subject to several limitations. First, due to the limited number of API calls available for closed-source LLMs, the datasets used for generating and detecting LLM-generated texts were constructed at a scale of 1,000 samples. As a result, the types and variety of texts involved in the analysis may not be fully comprehensive, potentially introducing bias. Additionally, because the generation of explanations requires manual annotation, which is time-consuming, only a random sample of 100 texts per dataset could be selected for evaluation. This sample size may lead to biases in the evaluation of LLM-generated explanations. Finally, given the rapid advancements in LLM technology, the detection and explanation capabilities of models are continually evolving. Therefore, it is crucial to periodically update our research focus and the models under study to ensure the results remain relevant and accurate.

\section*{Ethic Statements}

All experiments were conducted using publicly available LLMs and datasets. For the datasets we constructed for the work, no any personal or private information is included. All the three human annotators are co-authors, so an research ethics review was not considered necessary. More details on how we used the human annotators can be found in Appendix~\ref{appendix:annotation_guidelines}.

\bibliography{main}

\appendix

\section{Misattributions Examples}
\label{appendix:misattributions_examples}

To better illustrate the types of errors and their attributions, Table~\ref{tab:misattributions_examples} provides examples of common mistakes made by the models in classifying text origins.

\begin{table*}[h]
\centering
\small
\begin{tabular}{@{}p{\linewidth}@{}}
\toprule





\textbf{Example 2: Surface Features}

\textbf{Original Text}: ``The book were interesting. It show a deep exploration of human emotions and relationships. The author’s use of metaphors and descriptive language provided rich insights into the human psyche, but the writing style seemed awkward and unpolished in places.''

\textbf{LLM Judgment}: LLM-generated

\textbf{LLM Explanation}: ``The grammatical errors such as 'were' instead of 'was' and 'show' instead of 'shows' suggest that the text is AI-generated. Machines typically generate text with such mistakes.''\\

\midrule

\textbf{Example 3: Misattributed Logical or Emotional Elements}

\textbf{Original Text}: ``His words carried a strange contradiction—he spoke of hope but with an undeniable sadness. It was as though the optimism in his tone clashed with his deeply melancholic gaze, creating a complex emotional paradox.''

\textbf{LLM Judgment}: Human-written

\textbf{LLM Explanation}: ``The nuanced contradiction of hope and sadness reflects deeply human emotional complexity, which LLMs typically struggle to replicate authentically. Such layered emotional depth is often unique to human expression.''\\

\midrule






\textbf{Example 5: Fictitious Absence of Domain Knowledge}

\textbf{Original Text}: ``RNA interference plays a crucial role in gene regulation, allowing for targeted suppression of specific genetic sequences. This mechanism is being studied for its potential therapeutic applications in cancer treatment and genetic disorders.''

\textbf{LLM Judgment}: Human-written

\textbf{LLM Explanation}: ``The text lacks the level of technical depth and domain-specific knowledge typically found in professional scientific writing, suggesting that it is more likely to be written by a human author who is not an expert in the field.''\\






\bottomrule
\end{tabular}
\caption{Analysis of LLM vs. Human Writing Attribution Based on Various Features. The table categorizes examples where LLM-generated and human-written texts were incorrectly attributed or analyzed, providing explanations and analyses of these misattributions.}
\label{tab:misattributions_examples}
\end{table*}

\section{Supervised Fine-Tuning (SFT)}
\label{app:sft}

\subsection{Dataset Construction}

For each language model (e.g., GPT-4o, LLaMA3.3-8B), we first generate synthetic outputs and then perform self-detection using the same LLM. Based on whether the self-detection is correct, we apply the following rejection sampling procedure to construct our SFT datasets:

\begin{itemize}
  \item If the model correctly identifies its own output, we retain the sample.
  \item If the model misclassifies, but a stronger model (e.g., GPT-4o) can identify the source correctly and provide an explanation, we include this explanation-augmented sample.
  \item If all models fail to detect the text's origin correctly, the sample is discarded, as it offers limited training value.
\end{itemize}

This strategy ensures that our training data are high-quality and informative, containing either answers alone or answers with explanations. It helps the SFT model learn from both model-generated cues and external reasoning signals.

\subsection{Training Configuration}

Each SFT experiment is conducted using a balanced dataset of 10,000 LLM-generated and 10,000 human-written texts. We perform LoRA-based\cite{hu2021lora} fine-tuning for two epochs using four NVIDIA A100 80GB GPUs with FP16 precision. 

\begin{itemize}
  \item \textbf{Answer-only setting:} Each training session takes approximately 18 hours.
  \item \textbf{Answer + explanation setting:} Training requires around 28 hours due to longer input sequences and richer supervision.
\end{itemize}

We use a batch size that fully utilizes available GPU memory, a learning rate of 2e-5, and the AdamW optimizer. Model performance is evaluated after each epoch based on macro F1 score, and the best-performing checkpoint is selected.

\section{Reward Optimization with GRPO}
\label{app:rl}

\subsection{Dataset Construction}

For GRPO training, we construct a high-quality, moderately difficult dataset. We filter out both overly simple and excessively hard samples from the LLM-generated pool:

\begin{itemize}
  \item Samples that are correctly classified by all detectors are excluded as they lack training challenge.
  \item Samples that are misclassified by all detectors are removed because they may be inherently ambiguous.
\end{itemize}

From the remaining samples, we randomly select 5,000 LLM-generated texts and mix them with 5,000 human-written texts to form a balanced dataset for reward learning.

\subsection{Training Configuration}

GRPO training is conducted using the same infrastructure as SFT. We initialize the model either from a base Qwen2-7B checkpoint or from a previously SFT-trained model (cold-start setup). Training is performed using a PPO-style loop with KL-divergence regularization.

\begin{itemize}
  \item \textbf{Without SFT:} Training takes approximately 24 hours.
  \item \textbf{With SFT:} Training requires about 30 hours due to improved convergence and longer sequences.
\end{itemize}

All training is done with 4 NVIDIA A100 80GB GPUs, using FP16 precision, a reward model learning rate of 1e-5, a policy learning rate of 5e-6, and gradient accumulation for stability. The total training loop runs for 1 epoch with a replay buffer size of 10k examples.




\section{Annotation Guidelines}
\label{appendix:annotation_guidelines}

This appendix provides detailed instructions for the manual annotation tasks conducted in our study. The annotation process consists of two tasks: (1) a binary classification task to evaluate the accuracy of LLM-generated explanations, and (2) a ternary classification task where annotators evaluate both the correctness of the LLM's ternary classification judgments and the accuracy of its explanations based on known text sources.

\begin{table*}[h]
\centering
\small
\begin{tabular}{@{}p{\linewidth}@{}}
\toprule
\textbf{Task 1: Explanation Accuracy Evaluation}\\
Annotators will assess whether the explanation provided by the model correctly justifies its classification decision. Each explanation should be judged based on its logical consistency, relevance to the text, and whether it accurately reflects distinguishing features.\\

\textbf{Columns:}
\begin{itemize}
    \item \textbf{Text}: The text sample to be classified.
    \item \textbf{Detection Result}: The model’s classification of the text as ``LLM-generated'' or ``Human-generated.''
    \item \textbf{Model’s Explanation}: The explanation provided by the model for its classification decision.
    \item \textbf{Annotation}: Label the explanation as ``Accurate'' or ``Inaccurate'' based on its reasoning quality.
\end{itemize}\\

\textbf{Example:}\\
\textbf{Text}: ``In recent years, artificial intelligence has demonstrated remarkable progress, influencing numerous industries, including healthcare, finance, and creative writing. Many experts believe that this rapid advancement will continue to reshape the workforce and redefine human-machine collaboration.''\\
\textbf{Detection Result}: LLM-generated\\
\textbf{Model’s Explanation}: ``The highly structured argumentation and precise yet impersonal tone indicate that this text was likely machine-generated rather than composed by a human writer.''\\
\textbf{Annotation}: Inaccurate (While structured argumentation is common in LLM-generated text, human authors can also produce similarly structured and objective writing.)\\
\midrule

\textbf{Task 2: Ternary Classification with Explanation Evaluation}\\
Annotators will classify each text as ``LLM-generated,'' ``Human-generated,'' or ``Undecided,'' and evaluate whether the model’s explanation correctly justifies the classification. The ``Undecided'' label applies when the text lacks sufficient distinguishing features.\\

\textbf{Columns:}
\begin{itemize}
    \item \textbf{Text}: The text sample to be classified.
    \item \textbf{Detection Result}: LLM’ judgment of whether the text is ``LLM-generated,'' ``Human-generated,'' or ``Undecided.''
    \item \textbf{Model’s Explanation}: The explanation provided by the model.
    \item \textbf{Classification Annotation}: Label whether the model’s classification is ``Correct'' or ``Incorrect.''
    \item \textbf{Explanation Annotation}: Label the explanation as ``Accurate'' or ``Inaccurate'' based on its reasoning quality.
\end{itemize}\\

\textbf{Example:}\\
\textbf{Text}: ``Quantum mechanics, a field of physics that describes the behavior of particles at the atomic and subatomic levels, has led to groundbreaking discoveries such as quantum entanglement and superposition. These principles have paved the way for advancements in quantum computing and cryptography, revolutionizing modern technology.''\\
\textbf{Detection Result}: Undecided\\
\textbf{Model’s Explanation}: ``The text presents factual scientific content in a neutral tone, making it difficult to distinguish between human and machine authorship.''\\
\textbf{Classification Annotation}: Correct\\
\textbf{Explanation Annotation}: Accurate (The explanation correctly justifies why the text is indistinguishable.)\\
\bottomrule
\end{tabular}
\caption{Human Annotation Instructions}
\label{tab:annotation_instructions}
\end{table*}

\section{Examples of Undecided Text Categories}
\label{app:undecided_examples}

\subsection{Mixed Human-LLM Co-authored Texts}

This category includes texts collaboratively written by humans and large language models (LLMs), where stylistic or structural transitions reflect a shift in authorship. Such texts often begin with nuanced and context-sensitive human input and later transition to more uniform, templated, and encyclopedic LLM-generated content. These transitions are not always clearly marked, making authorship attribution difficult. Table~\ref{tab:mixed_texts} presents two representative examples.

\begin{table*}[h]
\centering
\small
\begin{tabular}{@{}p{\linewidth}@{}}
\toprule
\textbf{Example 1} \\

``In examining urban resilience frameworks, we find that grassroots organizations play a pivotal role in ensuring community adaptability. Interviews with local leaders in Jakarta revealed bottom-up innovation and resource sharing as key drivers of resilience. However, literature on climate-adaptive infrastructure increasingly emphasizes machine learning for real-time flood prediction and decentralized data systems for disaster response. A systematic review of recent publications demonstrates that hybrid models integrating sensor-based monitoring with socio-political data yield more actionable insights. These models offer scalable solutions for cities facing climate uncertainty, as evidenced by pilot projects in Southeast Asia and Latin America.'' \\

\midrule
\textbf{Example 2} \\

``The first wave of digital humanities emphasized textual digitization and basic metadata annotation, grounded in traditional philological practices. Scholars argued for methodological transparency and historical fidelity. In recent years, however, there has been a shift toward large-scale computational analysis. Transformer-based models are now trained on digitized archives to identify latent narrative patterns across centuries. This methodological turn, while powerful, raises questions about interpretability and disciplinary boundaries. Current debates focus on integrating humanistic inquiry with algorithmic inference in ways that preserve epistemic integrity.'' \\
\bottomrule
\end{tabular}
\caption{Examples of Mixed Human-LLM Co-authored Texts}
\label{tab:mixed_texts}
\end{table*}

\subsection{Inherently Ambiguous Single-source Texts}

This category includes texts produced entirely by either humans or LLMs, but whose stylistic and rhetorical features align with both sources. Such texts often exhibit emotionally neutral tones, fact-heavy content, and well-structured reasoning, making it difficult to distinguish their origin. Human-written texts may appear too polished, while LLM outputs may mimic human nuance. Table~\ref{tab:ambiguous_texts} illustrates two such ambiguous examples.

\begin{table*}[h]
\centering
\small
\begin{tabular}{@{}p{\linewidth}@{}}
\toprule
\textbf{Example 1} \\

``The monitor offers a 2560×1440 resolution, a 165Hz refresh rate, and a color accuracy rating of Delta E < 2. These specifications make it suitable for both competitive gaming and semi-professional design work. Its adjustable stand and blue-light reduction features enhance long-term usability. In testing, response times remained consistent across refresh rates, and input lag was minimal. While the built-in speakers are underwhelming, the overall design reflects thoughtful engineering. Prospective buyers should note that firmware updates may be required to access advanced color profiles.'' \\

\midrule
\textbf{Example 2} \\

``The exhibition explores post-colonial identity through mixed media installations that juxtapose industrial debris with archival imagery. Each piece is accompanied by minimal curatorial framing, allowing for open-ended engagement. The spatial arrangement avoids linear narratives, instead emphasizing fragmented temporality and layered symbolism. Visitor responses ranged from emotional introspection to conceptual confusion. Whether the ambiguity is intentional or a result of aesthetic overreach remains debatable, yet the collection undeniably provokes sustained reflection.'' \\
\bottomrule
\end{tabular}
\caption{Examples of Inherently Ambiguous Single-source Texts}
\label{tab:ambiguous_texts}
\end{table*}

\subsection{Fragile Indicator Cases}

This category includes texts that trigger LLM detectors based on subtle linguistic patterns or statistical anomalies. However, these indicators tend to be unstable. Minor changes in wording, paraphrasing, or model sampling parameters often cause the features to disappear. As a result, these texts demonstrate the brittleness of current detection methods. Table~\ref{tab:fragile_cases} presents two examples with unstable features that resist robust attribution.

\begin{table*}[h]
\centering
\small
\begin{tabular}{@{}p{\linewidth}@{}}
\toprule
\textbf{Example 1} \\

``The second quarter's economic indicators reflect a modest uptick in consumer confidence, despite lagging wage growth and persistent inflationary pressures. Analysts note that housing starts have stabilized, though regional variation remains high. Meanwhile, the energy sector showed unexpected resilience due to global supply chain recalibrations. Although many forecasts anticipated stagnation, revised models suggest a delayed soft landing. The Federal Reserve's policy stance continues to oscillate between caution and proactive intervention, with uncertainty surrounding long-term bond yields.'' \\

\midrule
\textbf{Example 2} \\

``In the novel’s final chapter, the protagonist revisits the childhood home, now transformed by decay and silence. The narrative perspective shifts from third-person to free indirect discourse, blurring the boundary between memory and present perception. Sentence rhythms slow, with nested subordinate clauses evoking emotional weight. Yet, this stylistic density is briefly interrupted by abrupt declaratives, mirroring the character’s psychological fragmentation. Such microstructural choices are atypical but could be easily altered in paraphrased variants, rendering authorship signals imperceptible to automated systems.'' \\
\bottomrule
\end{tabular}
\caption{Examples of Fragile Indicator Cases}
\label{tab:fragile_cases}
\end{table*}

\end{document}